%% file: main.tex
\documentclass[10pt,twocolumn,letterpaper]{article}

\usepackage[pagenumbers]{cvpr} 


\definecolor{cvprblue}{rgb}{0.21,0.49,0.74}
\usepackage[pagebackref,breaklinks,colorlinks,allcolors=cvprblue]{hyperref}
\usepackage{amssymb}  
\usepackage{pifont} 
\usepackage{gensymb}
\usepackage{multirow}
\usepackage{comment}
\usepackage[accsupp]{axessibility}

\usepackage{microtype}
\def\paperID{2813} 
\def\confName{CVPR}
\def\confYear{2025}

\title{ One2Any: One-Reference 6D Pose Estimation for Any Object}

\newcommand{\ourmodel}{One2Any}

\author{Mengya Liu$^{1}$ \quad Siyuan Li$^{1}$ \quad  Ajad Chhatkuli$^{2}$ \quad Prune Truong$^{3}$ \\
Luc Van Gool$^{1,2}$ \quad Federico Tombari$^{3,4}$ \\
\\
$^{1}$ ETH Zurich \quad $^{2}$ INSAIT, Sofia University  ``St. Kliment Ohridski" \quad $^{3}$ Google \quad $^{4}$ TUM
}

\begin{document}
\maketitle

\input{sec/0_abstract}    
\input{sec/1_intro}

\input{sec/2_related_works}

\input{sec/3_methods}

\input{sec/4_experiments}
\input{sec/5_conclusions}
\clearpage

\input{main.bbl}




\end{document}

%% file: sec/0_abstract.tex
\begin{abstract}
6D object pose estimation remains challenging for many applications due to dependencies on complete 3D models, multi-view images, or training limited to specific object categories. These requirements make generalization to novel objects difficult for which neither 3D models nor multi-view images may be available. To address this, we propose a novel method One2Any that estimates the relative 6-degrees of freedom (DOF) object pose using only a single reference-single query RGB-D image, without prior knowledge of its 3D model, multi-view data, or category constraints.
We treat object pose estimation as an encoding-decoding process: first, we obtain a comprehensive Reference Object Pose Embedding (ROPE) that encodes an object’s shape, orientation, and texture from a single reference view. Using this embedding, a U-Net-based pose decoding module produces Reference Object Coordinate (ROC) for new views, enabling fast and accurate pose estimation. This simple encoding-decoding framework allows our model to be trained on any pair-wise pose data, enabling large-scale training and demonstrating great scalability.
Experiments on multiple benchmark datasets demonstrate that our model generalizes well to novel objects, achieving state-of-the-art accuracy and robustness even rivaling methods that require multi-view or CAD inputs, at a fraction of compute. Code is available at \href{https://github.com/lmy1001/One2Any.git}{https://github.com/lmy1001/One2Any}.

\end{abstract}

%% file: sec/1_intro.tex
\section{Introduction}
\label{sec:intro}
6D object pose estimation is an important task in computer vision~\cite{nister2006scalable,philbin2007object,martinez2010moped}, due to its wide applicability in robotics, mixed reality and general scene understanding. However, the existing methods are still largely held back by generalizability and speed issues, as well as strict input requirements. Needless to say, these constraints are difficult to address together while keeping a good performance.

\begin{figure}
    \centering
    \includegraphics[width=1.0\linewidth]{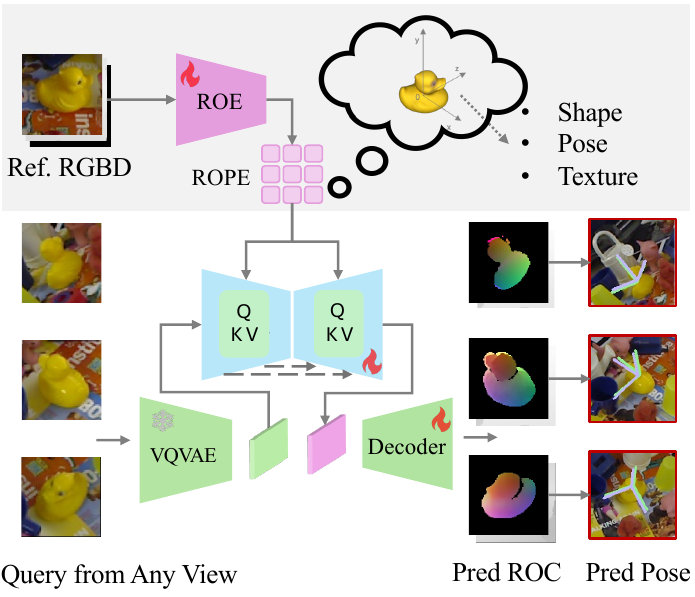}
    \vspace{-0.6cm}
    \caption{Given a single RGB-D image as a reference for an unseen object, our method estimates the pose of the object in a given query image, relative to the reference. The method first predicts a Reference Object Pose Embedding (ROPE) that encodes the object's texture, shape, and pose priors. During inference, each query RGB image is processed through a decoder to predict the Reference Object Coordinate (ROC) map and estimate the relative pose to the reference image. This approach effectively handles large viewpoint changes.}
    \label{fig:teaser}
\end{figure}

Existing works in 6D object pose estimation can be categorized based on the required inputs. In model-based approaches, a complete 3D model of the reference object is used at inference to support pose estimations~\cite{labbe2022megapose, nguyen2024gigapose,ausserlechner2024zs6d, lin2024sam, caraffa2025freeze, wang2021gdr}, while multi-view methods use many (8-200) reference images that indirectly encode the object’s 3D shape~\cite{he2022fs6d, park2020latentfusion, zhang2022relpose, liu2022gen6d}. Although effective, these methods are impractical in situations where multiple views or high-quality 3D models are unavailable, as is often the case for novel/unseen objects.
In contrast, direct absolute pose regression methods~\cite{xiang2017posecnn, kendall2015posenet} bypass extensive reference data through supervised learning but generally lack the ability to generalize to unseen objects.

Our goal is to estimate the pose of any object given only a single reference RGB-D view, a challenging setting for current methods. Multi-view approaches typically follow a “reconstruct-and-compare” pipeline, involving complex search, refinement, or bundle adjustment steps. The performance of these methods significantly drops when only sparse views are available, leading to poor reconstructions and inaccurate pose estimates. Alternatively, single-view methods based on 2D-2D correspondence~\cite{corsetti2024open, fan2024pope, zhang2022relpose} compute keypoint correspondences across views. However, such methods often struggle with textureless surfaces, occlusions, or large viewpoint gaps, making them unreliable in many practical scenarios (see Figure~\ref{fig:performance}).

To overcome the limitations of explicit reconstruction and 2D matching in sparse-view settings, we propose to learn a reference encoding-based conditioning. Recent advancements in 3D generation~\cite{liu2023zero, liu2024one} demonstrate that, when trained on large-scale datasets, latent diffusion models~\cite{rombach2022high,zhang2023adding} can be conditioned on pose, depth and additional information for image/3D generation. Building on this insight, we introduce a method built on latent diffusion architecture that learns robust, comprehensive conditioning to capture texture, shape, and orientation priors from a single reference view for pose estimation.

In this paper, we frame novel object pose estimation as a conditional pose generation problem: given a new view of an unseen object, we generate the object’s pose in a conditioned reference pose space. Our model comprises two branches: an instance encoding branch, which encodes a given RGB-D reference image into a Reference Object Pose Embedding (ROPE), and a Object Pose Decoding (OPD) branch, which takes the query image and ROPE to decode the object’s pose from any views as presented in Figure \ref{fig:teaser}. Instead of directly estimating rotation and translation, we introduce an intermediate dense representation suitable for our architecture.
Inspired by the Normalized Object Coordinate Space (NOCS)~\cite{ brachmann2014, taylor2012vitruvian}, which uses a canonical object pose to define 2D-3D correspondences for objects in a category, we relax the canonical frame requirement by defining Reference Object Coordinate (ROC) which presents a normalized object coordinate in the reference camera frame.

Our model builds on a pre-trained latent diffusion model~\cite{rombach2022high}, fine-tuning it to output ROC conditioned on the reference view with ROPE. With the generated ROC and target object depth, we compute the object’s pose efficiently through the Kabsch-Umeyama algorithm~\cite{umeyama1991least}. Furthermore, For faster inference, we bypass the diffusion process and run the U-Net in a feedforward fashion, enabling our method to achieve near real-time speed and operate significantly faster than existing approaches.

Extensive experiments on multiple object pose estimation benchmarks demonstrate that, even with only a single view, our method rivals approaches that rely on multiple images or CAD models, achieving state-of-the-art speed, accuracy and robustness.

%% file: sec/2_related_works.tex
\section{Related Works}
\label{sec:related_works}
Novel object pose estimation can be divided into three main setups based on the required inputs: model-based methods, multi-view methods, and single-view methods. Our approach falls under the third category, where we estimate object pose using only a single reference view and a single query view. We briefly review all three approaches below.

\noindent{}\textbf{Model-based novel object pose estimation. } CAD models are widely used in pose estimation tasks \cite{he2020pvn3d, he2021ffb6d, park2019pix2pose, labbe2022megapose, nguyen2024gigapose, chen2020learning, chen2021sgpa, he2022towards, irshad2022centersnap}. Instance-level methods rely on the same CAD model for training and testing, using 2D-3D correspondences \cite{park2019pix2pose}. Category-level approaches, like NOCS-based methods \cite{wang2019normalized, ikeda2024diffusionnocs, krishnan2025omninocs, cai2024ov9d}, learn a normalized canonical space for pose alignment within object categories, sometimes using category-level CAD models to handle shape variations \cite{chen2020learning, chen2021sgpa}. These approaches, however, are restricted to specific instances or categories and are less effective for novel objects. Recent works \cite{labbe2022megapose, nguyen2024gigapose, ausserlechner2024zs6d, lin2024sam, caraffa2025freeze, kirillov2023segment, ornek2025foundpose, wen2024foundationpose} address novel object pose estimation through “render and compare” strategies, generating multiple views from CAD models at test-time to estimate 2D-2D, 2D-3D, or 3D-3D correspondences. While effective, these methods are computationally intensive, and obtaining CAD models for unseen real-world objects remains challenging.

\begin{figure*}
    \centering\includegraphics[width=0.95\linewidth]{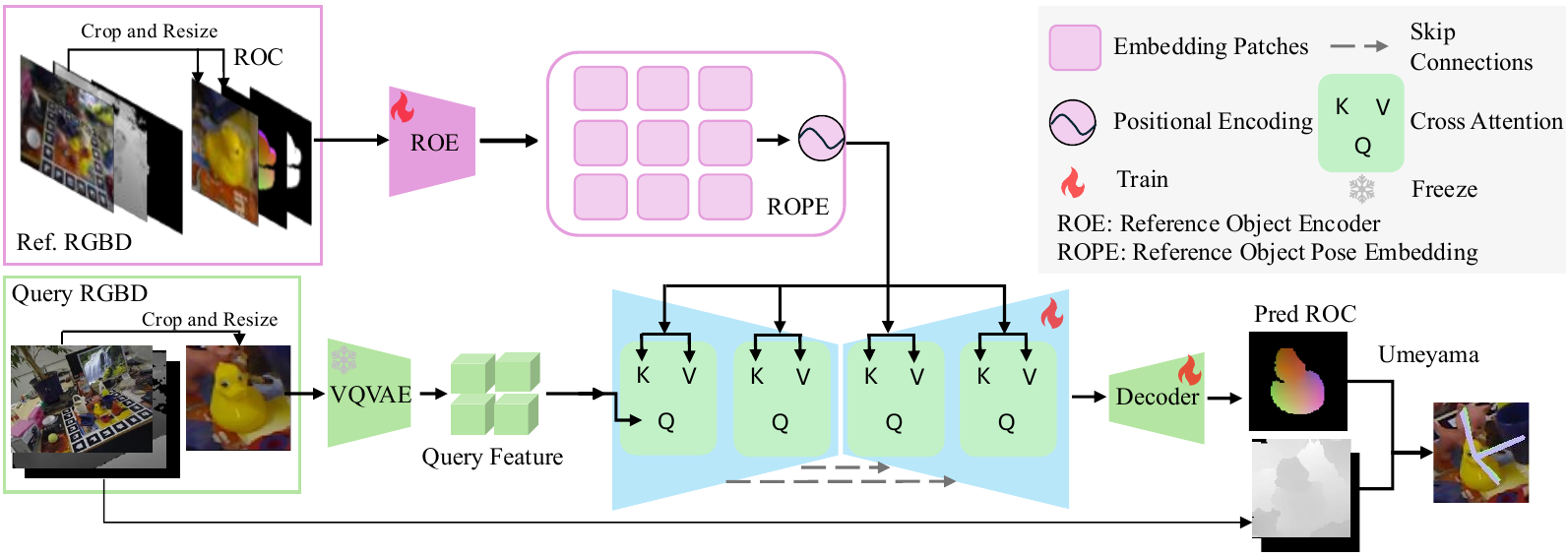}
    \vspace{-0.2cm}
    \caption{Network architecture. The network takes a reference RGB-D image as input and learns a Reference Object Pose Embedding (ROPE) through a Reference Object Encoder (ROE). This embedding is subsequently integrated with the query feature map, which is extracted using a pre-trained VQVAE model \cite{rombach2022high} with the query RGB image as input. We use the U-Net architecture for effective integrate the ROPE with the query feature with cross-attentions layers. The decoder is trained to predict the ROC map. The final pose estimation is computed using the Umeyama algorithm \cite{umeyama1991least}.}
    \label{fig:architecture}
\end{figure*}

\noindent{}\textbf{Multi-view novel object pose estimation. } Model-free approaches, which forgo CAD models, have gained popularity. They instead use multi-view images or a video sequence as supervision for novel object pose estimation. Methods like OnePose \cite{sun2022onepose}, OnePose++ \cite{he2022onepose++} and Gen6D \cite{liu2022gen6d} use RGB video sequences to do Structure-from-mMotion (SfM) to reconstruct the rough 3D unseen object and estimate the pose of the query. RelPose++ \cite{lin2023relpose++} and PoseDiffusion \cite{wang2023posediffusion} extend this approach to sparse view setups with bundle adjustment learning. However, these methods are computationally intensive, and performance drops notably when using fewer than 10 reference frames.  Some methods address these limitations with sparse views: FS6D \cite{he2022fs6d} uses dense matching by extracting RGBD prototypes from support views, but it still struggle with textureless objects due to the correspondence matching. While LatentFusion \cite{park2020latentfusion} builds a latent 3D object representation and renders more views based on it. It needs the sparse-view references to see the complete objects. FoundationPose \cite{wen2024foundationpose}, although CAD-model based, supports sparse views by constructing a neural object field for 3D model generation, yet its success heavily depends on informative and high quality multi views data of interesed object.

\noindent{}\textbf{Single-view novel object pose estimation. } Single-view pose estimation, which relies solely on a single reference view, is highly challenging. Methods in this category generally learn feature matching between two images to estimate relative pose.  For example, POPE~\cite{fan2024pope} uses DINOv2 features~\cite{oquab2023dinov2} for feature matching on novel objects. Oryon~\cite{corsetti2024open} integrates text prompts for object detection, fusing text and image features for improved matching, though it struggles with occlusions and textureless surfaces, leading to both translation and rotation errors. Horyon~\cite{corsetti2024high} enhances Oryon’s text-image fusion but faces similar limitations.  NOPE~\cite{nguyen2024nope} estimates a probability distribution over possible 3D poses for relative pose prediction but requires more than 300 potential relative poses at inference, leading to high computational costs and inaccurate discrete poses assumptions. Moreover, it does estimate translation. Our approach addresses these challenges by learning a continuous object latent space, avoiding the limitations of feature matching and enabling direct pose decoding.

Some point cloud registration methods solve the object pose estimation by detecting 3d keypoints \cite{li2019usip} or extracting feature descriptors \cite{huang2021predator, sun2021loftr, gumeli2023objectmatch}. These methods focus purely on the point cloud and fail to aggregate the 2D information from the image. Instead, our method is able to fully take advantage of the RGB image and the point cloud.



%% file: sec/3_methods.tex
\section{Method}
\label{sec:methods}

\subsection{Overview}
\label{sec:overview}
We formulate our problem as relative object pose estimation given a single reference RGB-D image and a query RGB-D image, without reliance on CAD models or multi-view images. 
Given one reference RGB-D image with RGB $A_{I}\in\mathbb{R}^{W \times H\times 3}$, depth $A_{D} \in \mathbb{R}^{W\times H}$, and target object mask $A_{M}\in\{0,1\}^{W\times H}$, our goal is to predict the pose $[\mathbf{R} | \mathbf{t}]$ of the target object in the query image input $Q_{I} \in \mathbb{R}^{W \times H\times 3}, Q_{D}\in \mathbb{R}^{W\times H}$, $Q_{M} \in \{0, 1\}^{W \times H}$, relative to the reference view. 
Our core idea is to learn a Reference Object Encoder (ROE), $f_A$, with parameters $\theta_A$, to embed the reference inputs $A = [A_{I}, A_D, A_M]$ into the Reference Object Pose Embedding (ROPE). By training $f_A$ over a large dataset, the embedding $\mathbf{e}_A$ conditions the Object Pose Decoding (OPD) module to generate the Reference Object Coordinate (ROC) map for the query image. We illustrate the network architecture in Figure. \ref{fig:architecture}. The OPD module contains an encoder-decoder architecture for extracting query image features and predicting the ROC map. This is further enhanced through integration with ROPE using a U-Net architecture. Here, we note $g_{\theta_{Q}}$ with parameters $\theta_Q$ as the OPD module, which takes a query image $Q$ as input to predict the output ROC: $Y_Q \in \mathbb{R}^{W\times H \times 3}$. We can then write the overall problem as,
\begin{equation}
\label{eq:model}
    Y_Q = g(Q, f_A(A;\theta_A); \theta_Q)
\end{equation}


\subsection{Reference Object Coordinate (ROC)}
\label{sec:nic_representation}
As described in Section \ref{sec:overview}, our objective is to estimate the relative pose $[\mathbf{R} | \mathbf{t}]$ between a single reference image $A$ and single query image $Q$. To that end, we propose to use a 2D-3D map, denoted as Reference Object Coordinate (ROC), inspired from NOCS~\cite{brachmann2014,wang2019normalized}. Unlike NOCS, our approach removes the canonical frame requirement and instead express the object coordinate directly within the reference camera frame. As a result, the ROC is defined solely by the reference frame, the object in the query image is transformed to align with reference frame and normalized to ROC space.

Although simple, the change has larger implications for both training and inference. NOCS, as defined, requires alignment of all objects in the same category to a single canonical space, ROC maps are much easier to generate, requiring only a pair of reference and query image, directly provide suitable representation for relative pose estimation.

\begin{figure}
    \centering
    \includegraphics[width=1.0\linewidth]{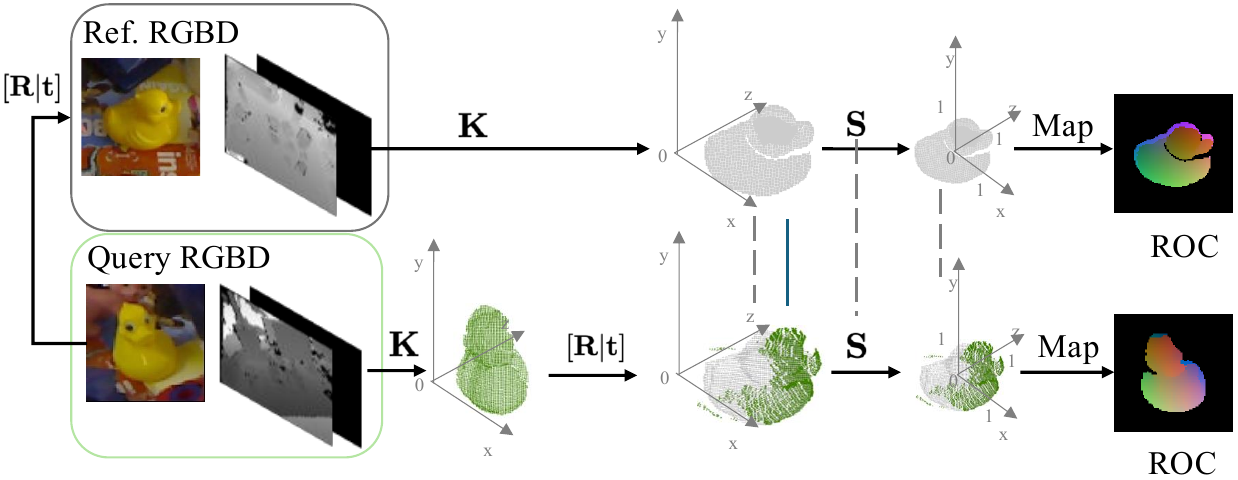}
    \vspace{-0.5cm}
    \caption{ROC representations given a reference RGB-D image and a query RGB-D image. The ROC space is initially defined by the reference frame, using the camera intrinsics  $\mathbf{K}$  and the scaling matrix  $\mathbf{S}$  to a normalized space. The query image is subsequently aligned to this space using the relative pose $ [\mathbf{R} | \mathbf{t}] $ and the scale matrix  $\mathbf{S}$. The ROC map is generated by mapping points in the ROC space to their corresponding 2D pixel locations and encoding the point positions as RGB values.}
    \label{fig:roc_representation}
\end{figure}

To construct the ROC from the reference view, we first obtain the partial point cloud $P_A$ of the reference object by backprojecting the pixel coordinates using depth:
\begin{equation}
    P_A = \mathbf{K}^{-1}  A_D[A_M=1]
\end{equation}
We then obtain ROC by applying scaling and shifting operations with the transformation $\mathbf{S}\in \mathbb{R}^{4\times4}$ operating on the homogeneous coordinates of $P_A$. 
\begin{equation}
    Y_A = \mathbf{S}  P_A    
\end{equation}
With a slight abuse of notation, we write the ROC which is a map: $A_{I} \to Y_A$, as $Y_A$. Similarly, in order to obtain ROC ground-truth for the query image with query view point cloud $P_Q$, we first obtain the query point cloud in the reference view using the ground-truth relative pose $[\mathbf{R}|\mathbf{t}]$ transformation, followed by the same scaling and shifting transformation $\mathbf{S}$. 
\begin{equation}
    Y_Q = \mathbf{S}\ ([\mathbf{R}|\mathbf{t}] \ P_Q )
\end{equation}
Figure \ref{fig:roc_representation} illustrates the process of constructing the ROC maps. The top part of the figure shows the generation of the ROC map $Y_A$ from the reference RGB-D image. The bottom part shows the generation of the ground-truth query ROC $Y_Q$. This approach establishes an object space based on the reference frame, which dynamically adapts with changes in the reference object and its pose.

\subsection{Reference Object Pose Embedding}
\label{sec:condition}
Pose prediction of a query RGB-D image, given only a single RGB-D reference image, presents unique challenges. Previous methods typically depend on keypoint feature matching \cite{nguyen2024nope, corsetti2024high, fan2024pope}, which fails when the visible regions in the images have low overlap or are heavily occluded. 
We propose an alternative approach that encodes the reference image into the Reference Object Pose Embedding (ROPE), allowing us to predict object poses effectively. 
Our objective is to train a Reference Object Encoder (ROE) that generates a comprehensive object representation in the latent space from a single RGB-D reference image. 
The ROPE representation then enables effective ROC map prediction on the reference camera space from any test image of the same object, thereby facilitating accurate pose estimation.

The ROE, denoted by $f(A; \theta_A)$ is designed to extract latent space encodings from a channel-wise concatenation: $A$ composed of the reference RGB image, reference ROC map, and the object mask. This encoding captures the texture and geometric information, as shown in Figure \ref{fig:architecture}. The encoder processes the inputs through three convolutional layers with residual connections \cite{he2016deep}, producing feature maps that are tokenized into patches with positional embedding \cite{alexey2020image}. This conditioned embedding effectively guides ROC map generation, maintaining fidelity to the reference data even in the presence of occlusion, as demonstrated in Section \ref{sec:ablation}.

\subsection{Object Pose Decoding with ROPE}
The Object Pose Decoding (OPD) module, $g(Q, f; \theta_Q)$ decodes the object pose in the query image based on ROPE derived from the reference image. With strong supervison from the representative ROPE, the OPD module predicts the object’s pose by generating the ROC map for the query image.

The OPD architecture includes an encoder-decoder architecture inspired from Stable Diffusion~\cite{rombach2022high}. To better aggregate the ROPE representation, we employ cross-attention layers to integrate information from the query image. Specifically, we use the pre-trained VQVAE model from \cite{rombach2022high} to fully extract the feature map of the query RGB image $Q_{I}$ and further improve the generalization.  We then finetune the corresponding UNet-like network~\cite{rombach2022high} conditioned on the representative ROPE via cross attention layers. We present a detailed diagram in Figure \ref{fig:architecture}.
The U-Net integrates the query feature $\mathcal{F}^{Q}$ and the reference embeding $\mathcal{F}^{A}$ (ROPE) through cross-attention. More precisely, in each cross-attention layer, the query feature $\mathcal{F}^{Q}$ interacts with the key $k \in \mathbb{R}^{m \times d_k} $ and value embeddings $ v \in \mathbb{R}^{m \times d_v} $ from $\mathcal{F}^{A}$.
\begin{equation}
    k = \mathcal{F}^A \times W_k, \quad  v = \mathcal{F}^A \times W_v ,  \quad   q = \mathcal{F}^Q \times W_q
\end{equation}
where $W$ denotes the weight matrix for each vector. And the cross-attention is applied between $q,k, v$,
\begin{equation}
    \text{Attention}(q, k, v) = \text{softmax}\left(\frac{q k^T}{\sqrt{d_k}}\right) v 
\end{equation}
$d_k$ is the dimensionality of the $k$ vector, used for scaling, it's applied in attention to ensure stable gradients. The final relative feature map $\mathcal{F}^{Q2A} $ is the implementation of multiple cross-attention layers through U-Net architecture \cite{rombach2022high}. The architecture allows the network to fully extract the pose-shape information embedded in ROPE.

The decoder then progressively refines and upsamples $\mathcal{F}^{Q2A}$, reconstructing it to the original image dimensions. This decoder, built with five convolutional layers featuring residual connections and bilinear upsampling, produces the ROC map $\hat{Y}_Q$, which accurately represents the query view coordinate in ROC space.

\noindent\textbf{ROC map loss.} Following NOCS~\cite{wang2019normalized}, we train the networks $\{f, g\}$ by supervising the ROC map prediction $\hat{Y}_{Q}$ with the ground-truth ROC map $Y_{Q}$ using a smooth L1-Loss~\cite{ren2016faster} as,
\begin{align}
\begin{split}
    \mathcal{L} &= \frac{1}{N}\sum_{i} \sum_{j}Q_M(i,j)E(i,j) \\
      c &= Y_{Q}(i,j) - \hat{Y}_{Q}(i,j)\\
    E(i,j) &= \begin{cases} 
      0.5 (c)^2 / \beta,  & (|c| < \beta) \\
      |c| - 0.5 \beta,  & \text{otherwise}\\
   \end{cases}
   \end{split}
\end{align}
$\beta$ is the smoothness threshold, set to 0.1. $Q_M(i,j)=1$ for object mask pixel positions and $0$ otherwise.

\subsection{Pose Estimation from ROC Map}
 The relative pose $[\mathbf{R} | \mathbf{t}]$ is computed by measuring the transformation between the predicted ROC $\hat{Y}_Q$ and the query view point cloud $P_Q$. Note that the reference camera pose is $[\mathbf{I}_3 |\mathbf{0}]$. First we shift and rescale the predicted ROC $\hat{Y}_Q$ to align to the reference camera coordinate, with the inverse of the scale matrix  $\mathbf{S}^{-1}$ from reference ROC. 
The point cloud thus obtained, ${\hat{P}_Q}^A$, denotes the query object in the reference camera frame. With the query point cloud $P_Q$, 
we use the Umeyama \cite{umeyama1991least} toobtain the pose prediction $[\hat{\mathbf{R}}|\hat{\mathbf{t}}]$ between the query frame and the reference frame. Mathematically, we can write it down as follows:
\begin{align}
\label{eq:poseprediction}
\begin{split}
        {\hat{P}_{Q}}^{A} &= \mathbf{S}^{-1}\hat{Y}_{Q}[Q_M=1] \\
    [\hat{\mathbf{R}}|\hat{\mathbf{t}}] &= \textbf{Umeyama}(P_Q, {\hat{P}_{Q}}^{A})
    \end{split}
\end{align}

%% file: sec/4_experiments.tex
\section{Experimental Results}
\label{sec:experiments}

\subsection{Training Details}
We train the reference object encoder and the query decoding module jointly as a unified framework. We train the model for 15 epochs with a batch size of 96 on 8 Nvidia RTX 4090 GPUs for around 2 weeks. We employ the AdamW optimizer  \cite{loshchilov2017decoupled}  with the learning rate of of 1e-4. Input images are pre-processed by cropping and resizing to $192 \times 192$ resolution. Our method is trained with ground-truth depth and mask images; however, in practical applications, it can also operate effectively with predicted masks like \cite{kirillov2023segment}, achieving comparable performance.

\subsection{Dataset}
\label{sec:dataset}
\textbf{Train set. } To robustly  enhance the model's generalization ability to real-world objects, we train our model on a synthetic dataset constructed in FoundationPose \cite{wen2024foundationpose}. 
FoundationPose sources its training assets from Objaverse \cite{deitke2023objaverse} and GSO \cite{downs2022google}, encompassing over 40,000 objects. 
These data include a wide range of daily-life objects in complex scenes. Each scene provides two distinct views; thus, we designate one view as a reference and the other as a query image. By cropping individual objects from each scene, this setup yields a dataset exceeding 2 million images.

To ensure the model is exposed to  various views for the object, we also incorporate a multi-view dataset into the train set.  The OO3D-9D \cite{cai2024ov9d} dataset, derived from OmniObject3D \cite{wu2023omniobject3d}, provides object-centric multi-view data across 5K objects and 216 categories. For our training, we utilize data from 200 categories consisting of 20K images. In ablation studies, due to computational constraints, we exclusively use this dataset for training and include an additional 10 unseen categories (approximately 230 objects) for testing.

\noindent{}\textbf{Test benchmark datasets. } To evaluate the generalization ability of our method in the real world, we test our method in the classic BOP benchmark dataset \cite{hodan2018bop}. This benchmark contains various real-world scenarios including occlusion, textureless objects, changing lights, significant view variations, and novel, unusual objects. Our evaluation includes the LINOMOD \cite{hinterstoisser2011multimodal}, YCB-Video \cite{xiang2017posecnn}, Toyota-Light \cite{hodan2018bop}, T-LESS \cite{hodan2017t} and LINEMOD-Occlusion \cite{hinterstoisser2011multimodal} datasets. Additionally, we assess our method on the widely-used Real275 test set \cite{wang2019normalized}, which includes 18 objects across 6 categories.

\begin{table}[]
    \centering
        \footnotesize
    \setlength{\tabcolsep}{0.8mm}
        \caption{Performance on Real275 \cite{wang2019normalized} and Toyota-Light \cite{hodan2018bop}. We compare our method with RGB and RGBD-based approaches when given only a single view as the reference. Each test set provides 2000 reference-query image pairs for evaluation.}
               \vspace{-0.2cm}
    \begin{tabular}{c|c | c c |c  c }
    \toprule
       \multirow{2}{*}{\textbf{Methods}}  & \multirow{2}{*}{\textbf{Modality}} & \multicolumn{2}{c|}{\textbf{Real275} \cite{wang2019normalized}} & \multicolumn{2}{c}{\textbf{Toyota-Light} \cite{hodan2018bop}} \\
      &  & AR  & ADD-0.1d & AR    & ADD-0.1d  \\
        \midrule
        PoseDiffusion \cite{wang2023posediffusion} & RGB & 9.2  & 0.8 & 7.8   & 1.2\\
        RelPose++ \cite{zhang2022relpose} & RGB  & 22.8 & 11.9 & 30.9  & 11.6 \\
        LatentFusion \cite{park2020latentfusion} & RGB &22.6  & 9.6 & 28.2  & 10.2 \\
         ObjectMatch \cite{gumeli2023objectmatch}& RGBD& 26.0 & 13.4 &9.8  & 5.4\\
         Oryon \cite{corsetti2024open}  & RGBD& 46.5 & 34.9 & 34.1 & 22.9\\
         \midrule
         One2Any(Ours) & RGBD & \textbf{54.9} &  \textbf{41.0} & \textbf{42.0} & \textbf{34.6} \\
         \bottomrule
    \end{tabular}
    \label{tab:oryon_dataset}
\end{table}

\begin{table*}[]
    \centering
    \footnotesize
   \setlength{\tabcolsep}{0.6mm}
       \caption{Performance on occluded YCB-Video \cite{xiang2017posecnn} dataset. We compare with point cloud registration methods, multi-view methods, and one-view methods. Predator \cite{huang2021predator}, LoFTR \cite{sun2021loftr} and FS6D \cite{he2022fs6d} are fine-tuned on the YCB-Video dataset. We evaluate ADD-S AUC and ADD AUC metrics. Results of multi-view methods are adopted from \cite{wen2024foundationpose}. For one-view methods, we provide the first image in the test set as the reference. We group objects into 5 categories, the best performance among multi-view methods and one-view methods are highlighted in bold. Detailed results can be found in the supplements.}
       \vspace{-0.2cm}
    \begin{tabular}{c| c c |c c | c c | c c || c c | c c | c c | c c}
    \toprule
        Methods  &\multicolumn{2}{c|}{\textbf{Predator} \cite{huang2021predator}} & \multicolumn{2}{c|}{\textbf{LoFTR} \cite{sun2021loftr}} &\multicolumn{2}{c|}{\textbf{FS6D} \cite{he2022fs6d}} & \multicolumn{2}{c||}{\textbf{FoundationPose} \cite{wen2024foundationpose}} &  \multicolumn{2}{c|}{\textbf{FoundationPose} \cite{wen2024foundationpose}} &  \multicolumn{2}{c|}{\textbf{Oryon} \cite{corsetti2024open}} &  \multicolumn{2}{c|}{\textbf{NOPE} \cite{nguyen2024nope}} & \multicolumn{2}{c}{\textbf{One2Any(Ours)}} \\
        Ref. Images & \multicolumn{2}{c|}{16} & \multicolumn{2}{c|}{16} & \multicolumn{2}{c|}{16} & \multicolumn{2}{c|}{16 - CAD}  & \multicolumn{2}{c|}{1 - CAD} & \multicolumn{2}{c|}{1} & \multicolumn{2}{c|}{1 + GT trans} & \multicolumn{2}{c}{1} \\
        \midrule
        metrics & ADD-S & ADD & ADD-S & ADD & ADD-S & ADD  &ADD-S & ADD & ADD-S & ADD & ADD-S & ADD & ADD-S & ADD & ADD-S & ADD \\
        \midrule

can &  73.6 & 29.23 & 76.08 & 51.07 & 91.9 & 50.0 & \textbf{97.2} & \textbf{85.2} & 90.8 & 81.5 & 21.3 & 12.4 & \textbf{95.6} & 32.9 & 94.4 & \textbf{84.9} \\
box  & 62.58 & 21.33 & 73.45 & 38.05 & 93.1 & 45.0 & \textbf{98.0} & \textbf{94.4} & 91.1 & 81.2 & 3.6 & 2.4 & 85.3 & 20.3 & \textbf{95.1} & \textbf{90.2}\\ 
bottles & 73.1& 23.62 & 52.02 & 24.6 & 87.7 & 39.1 & \textbf{97.0} & \textbf{90.5} & 90.0 & 73.3 & 4.3 & 3.7 & 89.3 & 26.7 & \textbf{93.0} & \textbf{81.9} \\ 
blocks & 74.85& 22.75 & 54.65 & 16.4 & 95.2 & 36.8 & \textbf{97.8} & \textbf{94.1} & \textbf{96.7} & 36.8 & 36.8 & 18.9 &  95.0 & 35.7 & 94.1 & \textbf{80.9}\\
others & 71.72 & 24.1 & 24.02 & 7.42 & 83.2 & 40.2 & \textbf{97.5} & \textbf{94.9} & 93.4 & 40.2 & 19.4 & 11.7 & 86.4 & 32.2 & \textbf{ 94.6} & \textbf{84.4}\\
\midrule
mean & 71.0 & 24.3 & 52.5 & 26.2 & 88.4 & 42.1 & \textbf{97.4} & \textbf{91.5} & 90.4 &76.1& 13.3 & 7.4 & 86.0 & 25.1 & \textbf{93.7} & \textbf{84.4}\\
         \bottomrule
    \end{tabular}
    \label{tab:ycbv_data}
\end{table*}

\subsection{Metrics}
Consistent with the evaluation protocol of FoundationPose \cite{wen2024foundationpose}, we assess pose estimation performance using the following metrics:
\begin{itemize}
\item Average Recall (AR) for Visible Surface Discrepancy (VSD), Maximum Symmetry-Aware Surface Distance (MSSD), and Maximum Symmetry-Aware Projection Distance (MSPD) metrics, as defined in the BOP\cite{hodan2018bop}.
\item ADD-0.1d: Recall of ADD errors within 0.1 of the object diameter (d), following the evaluation in \cite{corsetti2024open}.
\item ADD AUC and ADD-S AUC: Area Under the Curve (AUC) for ADD and ADD-S metrics, from \cite{xiang2017posecnn}.
\item $5\degree5cm, 10\degree5cm, 15\degree5cm$: We also evaluate average precision of object instances for errors  less than $5\degree5cm, 10\degree5cm, 15\degree5cm$ thresholds \cite{corsetti2024open, wang2019normalized}
.\end{itemize}

\subsection{Evaluation Procedure}
We primarily compare our method with single-view based approaches, including matching-based approaches Oryon \cite{corsetti2024open}, point cloud registration method ObjectMatch~\cite{gumeli2023objectmatch}, latent space pose learning method NOPE \cite{nguyen2024nope}, as well as multi-view based methods such as FoundationPose \cite{wen2024foundationpose}, LatentFusion \cite{park2020latentfusion}, and FS6D \cite{he2022fs6d}. For testing on the Real275 \cite{wang2019normalized} and Toyota-Light \cite{hodan2018bop} datasets, we follow the protocol established by Oryon, using 2000 image pairs in each test set. Baseline results are adopted from \cite{corsetti2024open}, with comparisons conducted assuming ground-truth masks are available. For testing on LINEMOD \cite{hinterstoisser2011multimodal} and YCB-Video \cite{xiang2017posecnn}, we adhere to the protocol in FoundationPose \cite{wen2024foundationpose}. Additionally, we evaluate a single-view reference setup, where \textbf{only the first view is used as the reference for the entire test set}. We directly test using the pretrained models provided by each method. For FoundationPose, we continue to generate the CAD model from the first view as supervisory input. Since NOPE only estimates rotation, we supplement its input with ground-truth translation values to ensure a fair comparison. Additional results are taken from \cite{wen2024foundationpose}.

\subsection{Pose Estimation Performance}
\textbf{Performance on real-world novel objects.} 
To evaluate our method’s effectiveness in handling real-world novel objects, we follow the protocol established by Oryon \cite{corsetti2024open} and conduct tests on the Real275 \cite{wang2019normalized} and Toyota-Light \cite{hodan2018bop} datasets. Real275 includes 18 objects across 6 categories, arranged in diverse indoor scenarios (e.g., on tables, floors) with a wide range of viewpoints. Toyota-Light introduces challenging lighting conditions for 21 distinct objects placed in various scenes. We test 2000 reference-query image pairs in each dataset. Results are shown in Table~\ref{tab:oryon_dataset}.
Our method significantly outperforms all alternative approaches in all metrics when only a single view is provided as a reference.

\noindent{}\textbf{Performance on occluded scenes. }
We evaluate our method on the occluded YCB-Video dataset \cite{xiang2017posecnn}. This dataset is particularly challenging, as some objects are largely occluded even in the first frame. Results for ADD-S AUC and ADD AUC are presented in Table \ref{tab:ycbv_data}.
Despite the occlusions in the YCB-Video dataset, our method outperforms all other one-reference approaches. In contrast, Oryon struggles to find reliable matches on occluded images, leading to poor results in both metrics. NOPE, on the other hand, faces challenges in learning pose distributions in the latent space when the reference view is occluded, resulting in poor relative rotation predictions. Even with ground-truth translations, NOPE underperforms on ADD-AUC, highlighting its limitations in handling real-world occlusions.

Surprisingly, our approach even outperforms all but one of the multi-view approaches, including FS6D, which has been fine-tuned specifically on YCB-Video dataset. Only FoundationPose-16view obtains better performance, as it leverages 16 views with ground-truth poses to generate a complete, high-quality CAD model. 

\begin{figure}
    \centering
    \includegraphics[width=1.0\linewidth]{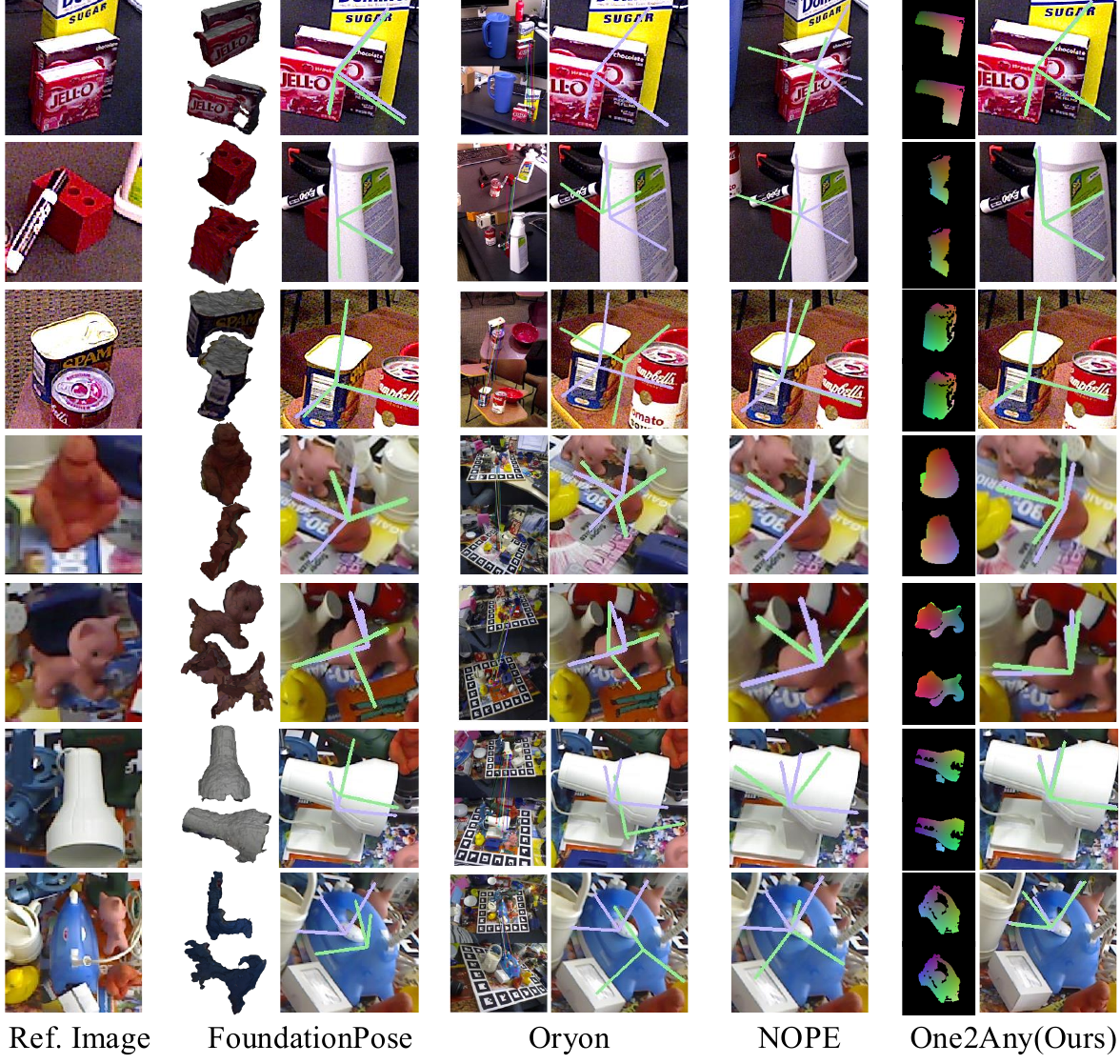}
    \vspace{-0.5cm}
    \caption{Qualitative results on YCB-Video \cite{xiang2017posecnn} and LINEMOD \cite{hinterstoisser2011multimodal} datasets. Predicted poses are displayed in green and ground-truth poses are in pink. We present FoundationPose \cite{wen2024foundationpose} with the generated CAD models from the reference image (the top is the view close to the reference image, and the bottom is the view close to the query image), and we display Oryon \cite{corsetti2024open} with the predicted correspondences. For our method, we also show the generated ROC map (bottom) compared with the GT ROC map (top).}
    \label{fig:performance}
\end{figure}

We also provide quantitative comparisons in Figure \ref{fig:performance}, with the first three rows depicting results from the YCB-Video dataset: (1) scenarios where the reference view is highly occluded, (2) where the query image is highly occluded, and (3) where the reference is occluded with a large view variation in query. 
As occlusions increase, only our method successfully predicts the relative pose of the query. 

\begin{table*}
\centering
    \footnotesize
    \setlength{\tabcolsep}{0.7mm}
    \caption{Performance on LINEMOD \cite{hinterstoisser2011multimodal} dataset with large view variations. In the top part of the table, we show results for multi-view approaches while in the bottom part, we compare single-view methods. Note that for all single-view methods, the first view is used as the reference. We report the recall of ADD-0.1d metric. Results of multi-view methods are taken from FoundationPose~\cite{wen2024foundationpose}. The best performance among multi-view methods and one-view methods are highlighted in bold.}
    \vspace{-0.2cm}
\begin{tabular}{l|c c| c c c c c c c c c c c c c|c}
\toprule
\textbf{Methods} & \textbf{Modality}  & \textbf{Ref. Images} & \textbf{ape} & \textbf{benchwise} & \textbf{cam} & \textbf{can} & \textbf{cat} &  \textbf{driller} & \textbf{duck} & \textbf{eggbox} & \textbf{glue} & \textbf{holepuncher} & \textbf{iron} & \textbf{lamp} & \textbf{phone} & \textbf{mean} \\
\midrule
OnePose \cite{sun2022onepose} & RGB & 200 & 11.8 & 92.6 & 88.1 & 77.2 & 47.9 & 74.5 & 34.2 & 71.3 & 37.5 & 54.9 & 89.2 & 87.6 & 60.6 & 63.6 \\
OnePose++ \cite{he2022onepose++} & RGB  & 200 & 31.2 & \textbf{97.3} & 88.0 & \textbf{89.8} & 70.4 & \textbf{92.5} & 42.3 & \textbf{99.7} & 48.0 & 69.7 & \textbf{97.4} & \textbf{97.8} & 76.0 & 76.9 \\
LatentFusion \cite{park2020latentfusion} & RGBD & 16 &  \textbf{88.0} & 92.4 & 74.4 & 88.8 & 94.5 & 91.7 & 68.1 & 96.3 & 49.4 & 82.1 & 74.6 & 94.7 & 91.5 & 83.6 \\
FS6D \cite{he2022fs6d} + ICP & RGBD & 16 & 78.0 & 88.5 & \textbf{91.0} & 89.5 & \textbf{97.5} & 92.0 & \textbf{75.5} & 99.5 & \textbf{99.5} & \textbf{96.0} & 87.5 & 97.0 &\textbf{97.5} & \textbf{91.5} \\
\midrule 
\midrule
FoundationPose \cite{wen2024foundationpose}& RGBD & 1-CAD & \textbf{36.5} & \textbf{55.5} & \textbf{84.2} & \textbf{71.7} &  65.3 &16.3  & \textbf{49.8} & 42.6 & 64.8 & \textbf{52.7} & 20.7 & 15.8 & 51.7 & 48.3\\
NOPE \cite{nguyen2024nope} & RGB & 1 + GT trans & 2.0 & 4.5 &  2.5 & 2.2 & 0.7 & 4.7& 0.5 &  \textbf{100.0} & 79.4 & 2.9  & 4.5 & 4.2 & 3.9 &16.3\\
Oryon \cite{corsetti2024open} &  RGBD & 1 & 1.2 & 1.3 & 3.9 & 0.8 & 12.7  & 8.5 & 0.8 & 63.2 & 18.4 & 1.6 & 0.6 &2.9 & 11.7 & 9.8 \\
\midrule
\ourmodel (Ours) & RGBD  & 1 & 33.1& 15.7 & 72.7 & 37.0 & \textbf{66.2}  & \textbf{68.2} & 35.8 & \textbf{100.0} & \textbf{99.9} &  42.0 & \textbf{28.2} & \textbf{31.9} & \textbf{53.2} & \textbf{52.6} \\

\bottomrule
\end{tabular}
    \label{tab:lm_data}
\end{table*}

\noindent{}\textbf{Performance on large view variations. }
We evaluate our method on the LINEMOD dataset \cite{hinterstoisser2011multimodal}, which contains various novel objects such as a toy ape, toy duck, and iron. The dataset includes both textureless and uncommon objects, presented in video sequences where the camera spans nearly 360 degrees around the scene. This setup poses a significant challenge for pose estimation, as using only the first view as a reference is particularly difficult under large view variations.

In Table~\ref{tab:lm_data}, we compare the recall of ADD-0.1d metric obtained by different approaches. Single-view method Oryon \cite{corsetti2024open} struggles to reliably extract matches for textureless objects, leading to complete failures on certain object categories, such as the toy ape and toy duck. NOPE \cite{nguyen2024nope} experiences a domain gap when handling highly novel objects, resulting in failures even when supplemented with ground-truth translations. 
In contrast, our method obtains the best performance in the one-reference setup. We are worse than the multi-view methods when they are provided with the multi-view images enough for a good 3d model reconstruction.
Notably, facing the large view variation with single reference image, our method is competitive with multi-view methods on some objects. 


We further qualitatively compare the four single-view methods in the bottom four rows of Figure \ref{fig:performance}.
While FoundationPose, Oryon and Nope are sensitive to incomplete CAD models, incorrect correspondences or novel objects respectively, our method successfully estimates poses even under large viewpoint changes (e.g., toy cat, lamp).
Moreover, FoundationPose's performance heavily depends on the quality of the CAD model. As illustrated with the iron example, when the CAD model is incomplete or inaccurate, its accuracy drastically drops even under small viewpoint variations. In contrast, our model is robust to a variety of configurations and setups.


\begin{table}[]
    \footnotesize
    \setlength{\tabcolsep}{0.6mm}
    \centering
        \caption{Performance for the pose tracking task on YCB-Video \cite{xiang2017posecnn} full video sequences. We compare with CAD-model based tracking methods. We give the first frame as a reference, and keep the reference unchanged during tracking. We report the average results over all objects, the best performance among multi-view methods and one-view methods are highlighted in bold. The detailed tracking performance for every object is in the supplements. }
        \vspace{-0.2cm}
    \begin{tabular}{c|ccc c|| c |c }
        \toprule
        
    \multirow{2}{*}{\textbf{Methods}}    & \textbf{Wuthrich} &\textbf{RGF} & \textbf{ICG} & \textbf{Foundation} & \textbf{Foundation} &  \textbf{One2Any} \\
    & \cite{wuthrich2013probabilistic}  &   \cite{issac2016depth} &  \cite{stoiber2022iterative} & \textbf{Pose} \cite{wen2024foundationpose}  & \textbf{Pose}  \cite{wen2024foundationpose}  & \textbf{(Ours)} \\
    \midrule
        Modality & CAD & CAD & CAD & 16-CAD & $1^{st}$-CAD & $1^{st}$\\
        \midrule
        ADD-S & 90.2 & 74.3 & 96.5 & \textbf{97.5}& 80.9 & \textbf{93.8}  \\
        ADD & 78.0 & 59.2 & 86.4 & \textbf{93.7} & 57.9 & \textbf{84.8}  \\
        
        \bottomrule
    \end{tabular}
    \label{tab:ycbv_tracking}
\end{table}

\subsection{Pose Tracking Performance}
In Table~\ref{tab:ycbv_tracking}, we present the pose tracking performance of our method on the YCB-Video dataset \cite{xiang2017posecnn} full video sequences. Following the evaluation protocol of FoundationPose \cite{wen2024foundationpose}, we compare our approach with CAD-model based methods Wuthrich \cite{wuthrich2013probabilistic}, RGF\cite{issac2016depth}, and ICG \cite{stoiber2022iterative}, that are specifically designed for pose tracking. Importantly, we use only the first frame as the reference for the entire tracking sequence. Our results demonstrate competitive performance relative to other model-based approaches. Additionally, when FoundationPose is restricted to using only the first view as the reference, we significantly outperform it. Our method also excels in per-frame pose estimation speed (see Table \ref{tab:inference_time}), an essential attribute for efficient pose tracking.

\subsection{Runtime Analysis}
Inference time is a critical metric for assessing the efficiency of methods in solving pose estimation tasks, particularly for pose tracking. We compare the runtime of our method with several state-of-the-art approaches on a single Nvidia 4090 GPU, as shown in Table \ref{tab:inference_time}. We compare  with CAD-model based, RGB based, and RGBD based methods. For CAD-model based methods, we compute the model ob-boarding time also in the whole process. Our method utilizes a feedforward encoder-decoder architecture, resulting in the fastest inference time. This efficiency is achieved by eliminating the need for reconstruction, rendering, or comparison steps required for pose search in most template-based methods, as well as omitting any refinement steps.

Other methods, such as MegaPose~\cite{labbe2022megapose} and GigaPose \cite{nguyen2024gigapose}, as CAD model based render-and-compare methods, exhibit longer runtime, even on high-performance GPUs. Note that, MegaPose and GigaPose cost 0.82 and 11.5s for templates generation respectively. FoundationPose, for instance, requires 1.4 seconds to generate pose hypotheses, and an additional 1.3 seconds for pose selection and refinement. NOPE \cite{nguyen2024nope} generates over 300 relative pose hypotheses for feature searching and comparison, making it highly time-intensive. Similarly, Oryon \cite{corsetti2024open} relies on feature matching, which is also slower compared to our approach.

\begin{table}[]
    \centering
        \footnotesize
        \setlength{\tabcolsep}{3.1mm}
        \caption{Inference time comparison. For CAD-model methods (MegaPose\cite{labbe2022megapose}, GigaPose \cite{nguyen2024gigapose}, FoundationPose \cite{wen2024foundationpose}), we also add the model rendering time. Our method is more than $10\times$ faster compared with other methods.}
        \vspace{-0.2cm}
    \begin{tabular}{c| c| c c }
         \toprule
        \textbf{Methods} & \textbf{Modality} & \textbf{GPU} &\textbf{Time (s)}  \\
       \midrule
        MegaPose \cite{labbe2022megapose} & CAD &  V100 GPU & 2.53 \\
        GigaPose \cite{nguyen2024gigapose} & CAD &  V100 GPU & 11.53 \\
        FoundationPose \cite{wen2024foundationpose} & CAD & RTX 4090 & 2.70 \\
        NOPE \cite{nguyen2024nope} & RGB & RTX 4090 & 20.15 \\
        Oryon \cite{corsetti2024open} & RGBD & RTX 4090 & 0.90\\
        \midrule
        One2Any(Ours) & RGBD & RTX4090 & \textbf{0.09} \\
         \bottomrule
    \end{tabular}
    \label{tab:inference_time}
\end{table}

\subsection{Ablation Study}
\label{sec:ablation}
Due to computational constraints, we conduct the ablation study using a subset of the training data as explained in Section \ref{sec:dataset}. Results are presented in Table \ref{tab:ablation_study}. 

\noindent{}\textbf{Ablation on ROC representation.} We first show the effectiveness of the ROC representation. We replace the decoder used to predict ROC map by a transformer-based rotation and translation head. 
The results, presented in Table \ref{tab:ablation_study} (first sub-table), show a significant drop in performance across all metrics when rotation and translation are predicted directly. This demonstrates the generalization attribute of the ROC representation in predicting pose. 

\begin{table}[t]
    \centering
        \footnotesize
        \setlength{\tabcolsep}{1.0mm}
            \caption{Ablation study on network design and reference, query feature representations. Each sub-table is related to one ablation study. The best performance is highlighted in bold.}
            \vspace{-0.2cm}
    \begin{tabular}{c| c c c c c }
    \toprule
        \textbf{Methods} & \textbf{ADD-S} & \textbf{ADD} & \textbf{5\degree5cm} & \textbf{10\degree5cm} & \textbf{10\degree10cm} \\
       \midrule
        rot head + trans head & 84.7 & 59.4&  2.4 & 8.9 & 9.2 \\
        ROC head & \textbf{91.2} & \textbf{78.6} & \textbf{15.5} & \textbf{40.5} & \textbf{40.7} \\
        \midrule
        \midrule
        Ref[RGB$\|$Mask] & 88.1& 73.2 & 8.2 & 24.9 & 25.2 \\
        Ref[RGB$\|$Depth$\|$Mask] & 88.7 & 74.8 & 8.8 & 28.9 & 29.2\\
        Ref[ROC$\|$Mask] & 90.5 & 76.8 & 13.3 & 34.1 & 34.2\\
        Ref[RGB$\|$ROC$\|$Mask] &\textbf{91.2} & \textbf{78.6} & \textbf{15.5} &\textbf{40.5} & \textbf{40.7} \\
        \midrule
        \midrule
        DINOv2~\cite{oquab2023dinov2} & \textbf{91.6} & \textbf{79.6} & \textbf{16.7} & \textbf{42.4} & \textbf{42.6} \\
        VQVAE~\cite{rombach2022high} & 91.2 & 78.6 &15.5 & 40.5 & 40.7\\
        \midrule
        \midrule 
        Query[RGB$\|$Depth] & 90.0 & 75.9 &8.9 & 38.2 & 38.3 \\
        Query[RGB] & \textbf{91.2} & \textbf{78.6} & \textbf{15.5} & \textbf{40.5} & \textbf{40.7}\\
         \bottomrule
    \end{tabular}
    \label{tab:ablation_study}
\end{table}

\noindent{}\textbf{Ablation on different reference representations.}  We measure the choice of our selected reference inputs in learning an optimal Reference Object Pose Embedding (ROPE) by experimenting with alternative input configurations. Our current reference input is a concatenation of reference[RGB$\|$ROC$\|$Mask]. We compare this with three variations: reference[RGB$\|$Mask], reference[ROC$\|$Mask], and reference[ RGB$\|$Depth$\|$Mask]. Results are presented in Table \ref{tab:ablation_study} (second sub-table).

The findings indicate that the ROC map is a critical component in the object latent representation. Using only the reference RGB image or depth concatenation for reference results in significantly lower performance compared to using only the reference ROC map. However, the ROC map alone does not convey texture information, which limits its effectiveness as a standalone representation. Overall, our reference representation, which incorporates both texture, geometric and pose information, provides a robust shape prior for the object.

\noindent{}\textbf{Ablation on different feature representations in the query decoding network.} We use a pretrained VQVAE~\cite{rombach2022high} as the primary feature extraction backbone. For comparison, we also experiment with pretrained DINOv2~\cite{oquab2023dinov2}. As shown in Table~\ref{tab:ablation_study} (third sub-table), DINOv2 achieves slightly better performance. This indicates that although VQVAE provides robust embeddings, other feature extractors can also be effectively integrated into our framework. Nevertheless, we select the pretrained frozen VQVAE for all experiments due to its stability and overall effectiveness. Additionally, we explore incorporating depth information into the query feature representation by concatenating RGB, depth, and mask inputs, denoted as Query[RGB$\|$Depth], which are fed through the small MLP layers followed by the feature extractor. From Table~\ref{tab:ablation_study} (last sub-table), we observe that including depth as an additional input degrades performance. This is likely because the pretrained VQVAE already provides highly effective RGB embeddings, and adding depth may introduce redundant or conflicting information.

%% file: sec/5_conclusions.tex
\section{Conclusions}
In this work, we introduced One2Any, a novel approach to 6D object pose estimation that requires only a single RGB-D reference view. By framing pose estimation as a conditional generation problem, our method leverages Reference Object Pose Embeddings (ROPE) and Reference Object Coordinates (ROC) to eliminate the reliance on CAD models or multi-view images. Through our modified latent diffusion architecture, we achieve state-of-the-art accuracy and speed on benchmark datasets, effectively generalizing to novel objects and challenging scenarios. Our method particularly excels in handling large viewpoint changes and occlusions that often challenge correspondence-based approaches. Furthermore, we achieve near real-time performance while maintaining high accuracy.